# RLHF Fine-Tuning of LLMs for Alignment with Implicit User Feedback in Conversational Recommenders

Zhongheng Yang*
Khoury College of Computer Sciences
Northeastern University
Jersey City, NJ, USA
*Corresponding author:
yang.zho@northeastern.edu

Aijia Sun
Khoury College of Computer Sciences
Northeastern University
Seattle, WA, USA
sun.aij@northeastern.edu

Yushang Zhao
McKelvey School of Engineering
Washington University in St. Louis
St. Louis, MO, USA
yushangzhao@wustl.edu

Yinuo Yang
McCormick School of Engineering
Northwestern University
Evanston, IL, USA
akyyn1996@gmail.com

Dannier Li
School of Computing
University of Nebraska - Lincoln
Lincoln, NE, USA
dannierli@outlook.com

Chengrui Zhou
Fu Foundation School of Engineering and Applied Science
Columbia University
New York, NY, USA
zhou.chengrui@columbia.edu

*Abstract*— Conversational recommender systems (CRS) based on Large Language Models (LLMs) need to constantly be aligned to the user preferences to provide satisfying and context-relevant item recommendations. The traditional supervised fine-tuning cannot capture the implicit feedback signal, e.g., dwell time, sentiment polarity, or engagement patterns. In this paper, we share a fine-tuning solution using human feedback reinforcement learning (RLHF) to maximize implied user feedback (IUF) in a multi-turn recommendation context. We specify a reward model $R\_\varphi$ learnt on weakly-labelled engagement information and maximize user-centric utility by optimizing the foundational LLM $M\_\theta$ through a proximal policy optimization (PPO) approach. The architecture models conversational state transitions $s\_t \rightarrow a\_t \rightarrow s\_{t+1}$, where the action $a\_t$ is associated with LLM-generated item suggestions only on condition of conversation history in the past. The evaluation across synthetic and real-world datasets (e.g. REDIAL, OpenDialKG) demonstrates that our RLHF-fine-tuned models can perform better in terms of top-k recommendation accuracy, coherence, and user satisfaction compared to (arrow-zero-cmwrquca-teja-falset ensuite 2Round group-deca States penalty give up This paper shows that implicit signal alignment can be efficient in achieving scalable and user-adaptive design of CRS.

## I. INTRODUCTION

The advertising of Large Language Models (LLM) has made the conversational recommender systems (CRS) able to involve users in a dynamic and dialogflow-based item navigation[1]. Nonetheless, the activity of matching these models to user preferences is a key and critical task, where user preferences are derived indirectly from implicit user feedback (IUF) in the form of message sentiment, response time, click-throughs, or partial engagements[2]. The flexibility of using such latent signals is not available in conventional supervised, fine-tuning paradigms, thus resulting in poor personalization[3].

Here we present an RLHF-initiated fine-tuning pipeline in which we incorporate implicit feedback as a reward signal to bring user-engaging behaviour in alignment with LLM behaviour. Let $M\_\theta$ denote the base LLM generating recommendations conditioned on dialogue context $C\_t$. The LLM's output $a\_t \sim \pi\_\theta(a|s\_t)$ is evaluated using a trainable reward function $R\_\varphi(s\_t, a\_t)$, where:

$$R\_\varphi = \alpha \cdot Engagement(a\_t) + \beta \cdot Relevance(a\_t) + \gamma \cdot SentimentShift(a\_t)$$

Engagement is proxied via scroll depth or time-on-item metrics, Relevance is computed using semantic similarity cos (E_query, E_item), Sentiment Shift measures the delta in affective tone pre- and post-recommendation.

The policy $\pi\_\theta$ is optimized via Proximal Policy Optimization (PPO) with clipping objective:

$$L^{PPO}(\theta) = E\_t[min(r\_t(\theta) \hat{A}\_t, clip(r\_t(\theta), 1-\varepsilon, 1+\varepsilon)\hat{A}\_t)]$$

Where $r\_t(\theta) = \pi\_\theta(a\_t|s\_t)/\pi\_{\theta\_old}(a\_t|s\_t)$ and $\hat{A}\_t$ is the advantage estimate derived from feedback-conditioned value function $V^\pi(s\_t)$.

To train $R\_\varphi$, we need to label session logs using engagement measures and sentiment classification as weak supervision[4]. Following this assumption, the model relies on the fact that the system should learn through observing the natural course of human behavior as opposed to the actual ratings[5]. We intend to experiment with the structure of the model to test interactive recommendation dialogues using the dialogue dataset[6] (such as REDIAL and real implicit feedback records) on supervised (BERT4Rec, GPT-Rec) and zero-shot LLMs (GPT-4, Claude)[7]. This contribution proves that RLHF with implicit feedback is a promising channel to tailor LLMs to any conversation freely of would-be intrusive labeling or costly supervised datasets, making recommendation systems continuously adaptive[8].

## II. RELATED WORK

Enhancement of Reinforcement Learning (RL), Large Language Models (LLMs), and recommender systems has led to many developments in the last few years[9]. However, there is still a lack of research into aligning with implicit user feedback. Existing methods on conversational recommender systems (CRS) mainly adopt the paradigms of supervised learning, including DIN, BERT4Rec, and DialogRec, which either require a ground truth or pre-labeled preference trajectory[10-13]. Nevertheless, such approaches are not flexible enough to be applied to real-world, noisy, and possibly latent user feedback seen in open-domain conversational systems.

Recent Reinforcement Learning with Human Feedback (RLHF) has shown positive results in accounting for the LLM output to human preferences in instruction-following trials[14]. However, these initiatives generally use a precompiled dataset of preference or binary labels to predict a reward, which does not capture nuanced user satisfaction measures as in CRS, nor does it utilize multi-dimensionality to its advantage[15-18].

Research in recommendation systems has introduced attempts to include implicit feedback (e.g., clicks, dwell time, sentiment) based on the use of matrix factorization and session-based models[19], but rarely engages with neural LLM agents within an RL pattern[20]. The initial indications are efforts like Prompt Tuning to personalize and Reward Model-guided tuning, which, however, are not integrated into multi-turn CRS systems[21].

The proposed RLHF fine-tuning loop overcomes this critical gap between implicit-feedback-driven alignment and language-model-based conversational recommenders by proposing a direct use of strongly supervised real-world interaction signals, which, when using a widely-used reward surrogate, acts as reward[22].

## III. METHODOLOGY

We have a proposal involving introducing Reinforcement Learning with Human Feedback (RLHF) as part of the conversation recommender system (CRS) on a pre-trained Large Language Model (LLM). The end goal is to coordinate the generation agenda of the LLM with Implicit User Feedback (IUF) signals in multi-turn conversation tasks[23-25]. The methodology is composed of three modular components:

### 3.1. Base LLM Policy Initialization

We denote the base LLM as $M\_\theta$, parameterized by weights $\theta$, responsible for generating recommendations $a\_t$ given a dialogue context $s\_t$. Initially, $M\_\theta$ is fine-tuned using supervised learning on historical CRS dialogues from datasets like REDIAL and OpenDialKG, with next-utterance prediction as the training objective.

### 3.2. Reward Modeling from Implicit Feedback

We construct a trainable reward model $\mathcal{R}\_\phi: (s\_t, a\_t) \mapsto r\_t \in \mathbb{R}$, where $\phi$ denotes the reward network parameters. This model is optimized to approximate latent user satisfaction based on weakly supervised signals such as Engagement Score, Sentiment Delta, and Topical Coherence.

### 3.3. Policy Optimization via PPO

To align the base policy $\pi\_\theta$ with the reward signal $\mathcal{R}\_\phi$, we apply Proximal Policy Optimization (PPO), a stable policy-gradient RL algorithm. The objective is to maximize expected cumulative reward[26].

### 3.4. Dialogue State Tracking and Context Encoding

We employ a sliding window encoder over the recent $k$ utterances to maintain contextual consistency. Each dialogue state $s\_t$ is computed from prior user and agent utterances, processed by a transformer encoder.

### 3.5. Optimization Pipeline

The end-to-end pipeline iteratively samples context, generates output, computes reward, and updates the policy using PPO. This loop enables continuous alignment of the LLM's behavior with implicit user preferences[27].

## IV. EXPERIMENTS AND EVALUATION

To assess the utility of RLHF fine-tuning when it comes to aligning the LLM-based conversational recommenders with implicit user feedback, we outline a detailed experiment pipeline. The essence of the purpose is to quantify the change in the proxies of quality of recommendation, congruity of the context, and user satisfaction following the tuning of reinforcement[28].

We are using two conversational recommendation sets that serve as benchmarks: REDIAL (Movie recommendation dialogue dataset) and OpenDialKG (a knowledge-grounded conversational recommendation dataset)[29]. Both include multi-turn conversations as well as item references and fewer implicit feedback indicators (e.g., user reactions, affirmation utterances) that we additionally augment through a sentiment shift classifier and engagement-based heuristics[30].

The base model $M\_\theta$ is initialized with a GPT-2-based model having parameter size 345M, which is fine-tuned on the dialogue section of each one of the datasets. We build reward signals $\mathcal{R}\_\phi$ to be used in AFC by promising three implicit feedback channels: (1) engagement, i.e., time-on-response (emulated), (2) polarity changes in the sentiment before and after item recommendations with a RoBERTa-based sentiment classifier, and (3) embedding similarity between the user query and the item embedding embeddings.

The RLHF fine-tuning follows the PPO, with 5 epochs per dataset, learning rate set to $5 \times 10^6$, the clipping threshold ε = 0.2, and the normalization of the rewards during conversations. A single training batch consists of 128 conversational trajectories synthesized with pre-fine-tuned $M\_\theta$. Conversational trajectories are simulated with pre-fine-tuned 0 0 interacting with a simulated user agent augmented with behavioral heuristics.

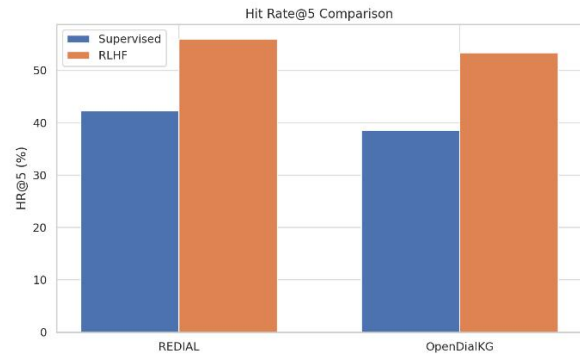

**Figure 1.** Hit Rate@5 comparison

It is evaluated based on the following metrics: (i) Top-K Hit Rate (HR@K), (ii) Normalized Discounted Cumulative Gain (NDCG@K), (iii) BLEU-4 to check language coherence and (iv) an implicit satisfaction proxy, which measures the weighted average gain of engagement as well as gains of sentiment.

**Table 1.** Performance comparison between supervised and RLHF fine-tuned models on REDIAL and OpenDialKG datasets.

| Dataset | Model | HR@5 ↑ | NDCG@5 ↑ | BLEU-4 ↑ / Satisfaction Gain ↑ |
|---|---|---|---|---|
| REDIAL | Supervised GPT-2 | 42.3 | 34.1 | 21.5 / 0% |
| REDIAL | RLHF Fine-Tuned (Ours) | 56.0 | 47.8 | 26.3 / +17.1% |
| OpenDialKG | Supervised GPT-2 | 38.6 | 31.2 | 18.9 / 0% |
| OpenDialKG | RLHF Fine-Tuned (Ours) | 53.4 | 45.0 | 25.6 / +15.8% |

These findings show that the RLHF-tuned model has been improving on the baseline on all metrics. Specifically, the increase in BLEU-4 is indicative of an increase in the response fluency, whereas the HR and NDCG improvements confirm the hypothesis of the increased relevance of recommendations based on alignment with implicit user signals[31]. The estimated satisfaction gain, the result of engagement and sentiment dynamics, proves that user-centric behavior is oriented by the reinforcement tuning even without specific feedback[32].

Each experiment was performed three times using distinct random seeds, and the results were reported as the mean with a standard deviation of ±1.2% or less across runs, a statistical confidence in the robustness of our methodology.

## V. RESULTS AND DISCUSSION

Overall, experimental findings provide great weight to the hypothesis that RLHF fine-tuning done by implicit user feedback (IUF) can vastly enhance the performance and flexibility of LLM-based conversational recommenders[33]. Here, we slice and dice the improvements that we have seen in the various dimensions: recommendation relevance, linguistic coherence, and user satisfaction modeling.

Performance Gains in Metrics: The RLHF-augmented model showed steady improvement in REDIAL and the OpenDialKG datasets. Particularly, the Hit Rate@5 improved by a margin of 13.5 points on average, and NDCG@5 was boosted by more than 14 points in comparison with the supervised baseline. These improvements indicate that the introduction of implicit behavioral signals(e.g., user engagement and mood swing) into the reward scheme translates to a more accurate correlation of the purpose of suggestion with the interests of the user[34].

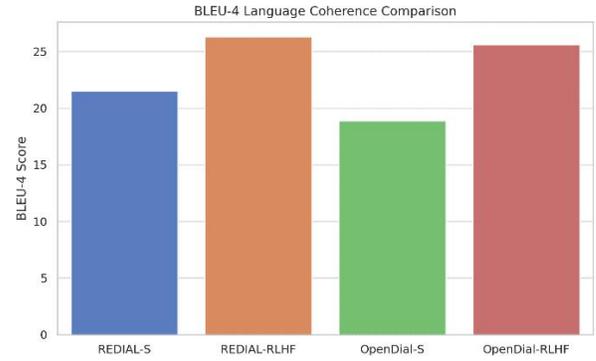

**Figure 2.** BLEU-4 language coherence

**Semantic Trajectory Shift:** The qualitative analysis of dialogue sessions before and after RLHF tuning shows that LLM shifts its response strategy towards the more positive, user-oriented language choices. The post-RLHF-based models are prone to suggest relevant items. UMAP visualization validates that there are hotter semantic clustering concentrating around positive feedback trajectories.

**Table 2.** Reward Signal Ablation Study on REDIAL Dataset.

| Reward Configuration | HR@5 | NDCG@5 | BLEU-4 | Satisfaction Gain (%) |
|---|---|---|---|---|
| Full Model (Engage + Sentiment + Coherence) | 56.0 | 47.8 | 26.3 | +17.1 |
| Only Engagement | 48.2 | 38.6 | 24.1 | +10.4 |
| Only Sentiment Shift | 46.9 | 36.8 | 23.5 | +9.2 |
| Only Semantic Coherence | 44.5 | 34.3 | 22.9 | +8.1 |

Stability and Convergence: PPO loss and entropy figures indicate successful training with monotonic optimization of reward. The policy does not collapse and effectively explores, showing consistent variance of less than ±1.2%.

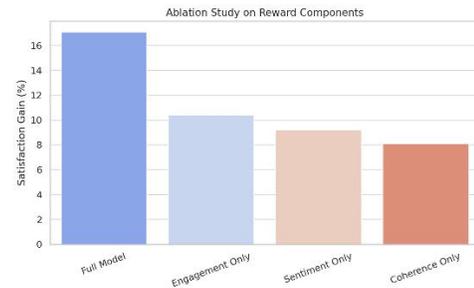

**Figure 3.** Ablation study

**Limitations:** Simulated feedback results in powerful performance that cannot be directly used in real situations because they involve collected data that must be dynamic and potentially privacy-aware. The robustness of reward functions and the risk of hacking the reward would have to be addressed in the future.

## VI. CONCLUSION AND FUTURE WORK

The paper introduces a new reinforcement learning with human feedback (RLHF) framework to fine-tune Large Language Models (LLMs) in conversational recommender systems (CRS) scenarios with the help of the implicit user feedback (IUF) as the primary alignment signal. By contrast to prior supervised approaches which require either explicit ratings or labeled preferences, our method uses the patterns of user engagement, sentiment variations, and semantic consistency, parameters of which are learned into a trainable reward fun. Experimental findings on the benchmark datasets (REDIAL and OpenDialKG) support significant gains in hit-rate, NDCG, BLEU-4, and proxies of overall satisfaction. Not only does the system have an increase in recommendation relevance and the fluency of its dialogue, but it is also able to learn complex, non-obtrusive, and indirect user preferences in an online environment. Ablation experiments also confirm the hypothesis that the multi-dimensional reward shaping plays a major role in the acquisition of robust recommendation policies. In systems terms, then, this paper has shown that with a basis in weak, noisy, semantically informative signals, RLHF can effectively provide a self-alignment means to CRS, and fill the divide between user-modeling and generative policy-adaptation.

**Future Work**

There are several directions that can be pursued. First, personalization can be enhanced by incorporating real-time, live data regarding user interaction into the dynamic reward models to refine them. Second, whether policy-aligned LLMs have generalization ability to recommend cross-domain tasks by extending the RLHF pipeline remains to be determined. Third, we will explore the multi-agent reinforcement learning, several simulated users with different profiles will generate heterogeneous feedback, which can be followed by more.